\documentclass{article}

\usepackage{PRIMEarxiv}

\usepackage[utf8]{inputenc} %
\usepackage[T1]{fontenc}    %
\usepackage{hyperref}       %
\usepackage{url}            %
\usepackage{booktabs}       %
\usepackage{amsfonts}       %
\usepackage{nicefrac}       %
\usepackage{microtype}      %
\usepackage{lipsum}
\usepackage{fancyhdr}       %
\usepackage{graphicx}       %
\graphicspath{{media/}}     %

\usepackage{amsmath}
\usepackage{tabularx}
\usepackage{todonotes}
\usepackage{multirow}
\usepackage{natbib}[authoryear]
\usepackage{makecell}

\title{CeRA: Breaking the Linear Ceiling of Low-Rank Adaptation with Non-linearity Retained at Inference
}

\author{
  Hung-Hsuan Chen \\
  Computer Science and Information Engineering \\
  National Central University \\
  Taoyuan, Taiwan \\
  \texttt{hhchen1105@acm.org}
}

\begin{document}
\maketitle

\begin{abstract}

  Low-Rank Adaptation (LoRA) dominates parameter-efficient fine-tuning (PEFT). However, it faces a ``linear ceiling'': increasing the rank yields diminishing returns in expressive capacity due to linear constraints. We introduce CeRA (Capacity-enhanced Rank Adaptation), a weight-level parallel adapter that injects SiLU gating and dropout to induce non-linearity during inference, thereby placing it in a different function class from adapters whose non-linearity exists during training and collapses to an affine map at inference time. On both the basic arithmetic (GSM8K) and the complex MATH benchmark, CeRA is markedly more parameter-efficient. Across a full rank $\times$ learning rate sweep, CeRA at rank 64 achieves the highest MATH pass@1 of any configuration in the grid (23.6\%), matching or exceeding both a rank-512 LoRA (22.4\%) and DoRA (19.8\%) while using only 1/8 of the parameter budget. With the rank and learning rate fixed, CeRA equals or outperforms LoRA in 10 of 12 matched settings. Spectrally, CeRA's learned updates utilize the singular-value spectrum more broadly than linear adapters, which exhibit rank collapse at high rank, although a scale-matched control shows that this difference stems mostly from output scale and partially from non-linearity. Additionally, dropout appears to contribute to regularization rather than rank expansion. We release the code for reproducibility.\footnote{\url{https://github.com/hhchen1105/cera}}

\end{abstract}


\section{Introduction} \label{sec:intro}

Parameter-Efficient Fine-Tuning (PEFT) is the de facto standard for fine-tuning Large Language Models (LLMs). Among various techniques, Low-Rank Adaptation (LoRA)~\citep{hu2022lora} is widely used. Its design relies on the mergeability assumption: weight updates must be linear ($\Delta W = BA$) to allow merging with the base model for zero-latency inference.

We challenge this linear constraint. Variants that attempt to refine LoRA, such as weight decomposition in DoRA~\citep{liu2024dora} or adaptive rank allocation~\citep{zhang2023adalora}, primarily focus on optimizing the learning dynamics of the linear subspace but leave the hypothesis space unchanged. These methods remain bounded by the expressivity limits of linear transformations, creating a ceiling for reasoning-intensive tasks. In our experiments, a high-rank LoRA ($r=512$) performs similarly to its low-rank counterpart ($r=64$); since overparameterization typically eases optimization even in linear models~\citep{chen2020accelerating}, this suggests that the bottleneck may be the structural rigidity of linearity itself rather than the parameter count or optimization difficulty.

To overcome this ceiling, we introduce CeRA (Capacity-enhanced Rank Adaptation). CeRA shifts from linear space optimization to nonlinear. By injecting SiLU gating and dropout into a \emph{weight}-level parallel adapter, CeRA provides the high-dimensional expressivity required for complex reasoning. Furthermore, in multi-tenant serving where many adapters share a single frozen base model and are kept unmerged---in the cloud (e.g., S-LoRA~\citep{sheng2023slora}, Punica~\citep{chen2024punica}) and on edge devices (e.g., EdgeLoRA~\citep{shen2025edgelora})---the latency cost of non-linearity is modest rather than prohibitive (details in Section~\ref{subsec:complexity}).

Our contributions are threefold:
\begin{itemize}
    \item We propose CeRA, a fine-grained, weight-level parallel adapter that integrates nonlinear gating to capture complex functional updates beyond linear approximations.
    \item We evaluate downstream exact-match accuracy and highlight the role of task complexity. On the challenging MATH dataset, CeRA ($r=64$) matches or exceeds high-rank LoRA ($r=512$) and DoRA, showing that capacity expansion improves reasoning without inflating the parameter budget.
    \item Using Singular Value Decomposition (SVD), we show that CeRA's learned updates spread energy into the tail of the singular value spectrum where linear methods collapse, and---via a scale-matched control experiment (Appendix~\ref{app:er_control})---that much of this difference stems from output scale, so we treat spectral expansion as a descriptive diagnostic rather than the mechanism behind the accuracy gains.
\end{itemize}

\section{CeRA: Capacity-enhanced Rank Adaptation} \label{sec:method}

\subsection{Preliminaries: The Linear Confinement}

For a pre-trained weight matrix $W_0 \in \mathbb{R}^{d \times k}$, LoRA constrains the update $\Delta W$ to a low-rank decomposition $BA$, where $B \in \mathbb{R}^{d \times r}$, $A \in \mathbb{R}^{r \times k}$, and $r \ll \min(d, k)$. The forward pass is:
\begin{equation}
    h = W_0 x + \frac{\alpha}{r} BA x,
\end{equation}
where $\alpha$ is a hyperparameter.

This formulation allows $\Delta W$ to merge into $W_0$. However, we hypothesize that this constraint contributes to the rank under-utilization observed in complex tasks.

\subsection{The CeRA Architecture}

CeRA relaxes this linear constraint while retaining the parallel bottleneck structure to maintain parameter efficiency. Formally, CeRA defines the forward pass as:
\begin{equation}
    h = W_0 x + s \cdot B\,\mathcal{D}(\sigma(A x)),
\end{equation}
where $A \in \mathbb{R}^{r \times k}$ projects the input to a latent dimension $r$, $\sigma(\cdot)$ is SiLU, $\mathcal{D}(\cdot)$ is dropout, $B \in \mathbb{R}^{d \times r}$ projects back to the output dimension, and $s$ is a scaling scalar.

\paragraph{Weight-Level Granularity and Non-Linearity.}
Unlike module-level parallel adapters~\citep{he2021towards, zhu2021counter} that process the aggregate output of an entire attention block, CeRA operates at the weight level. By injecting updates into the internal query ($W_q$) and value ($W_v$) projections, CeRA alters the attention mechanism's internal feature dynamics. The SiLU activation enables the adapter to suppress noise or amplify feature directions. Empirically, it is the component that broadens the adapter's effective rank (Section~\ref{subsec:ablation}). Dropout $\mathcal{D}$ regularizes the bottleneck latent space and improves predictive performance, consistent with prior findings that structured edge-dropping regularization improves robustness~\citep{yang2025dynamic}.

\section{Experiments} \label{sec:experiments}

We use Llama-3.1-8B as the frozen backbone for the main experiments and vary only the adapter architecture. We train the models in \texttt{bfloat16} precision with AdamW optimizer.\footnote{We verify the effectiveness of CeRA across model scales (1B/3B/8B) in Appendix~\ref{app:model-scale}.} For LoRA and DoRA, we use a fixed scaling hyperparameter $\alpha = 32$ (the common PEFT default). CeRA uses a fixed scale $s = 1$ throughout. For each method and rank, we select the learning rate based on MATH pass@1 (the full sweep is in Appendix~\ref{app:hyperparameters}), which partially mitigates the coupling between the adapter scale and the learning rate.

We run two \emph{separate} fine-tuning experiments, one per training dataset, to investigate capacity scaling and domain robustness: SlimOrca~\citep{slimorca} ($\sim$300k GPT-4-augmented instruction pairs emphasizing Chain-of-Thought) and MathInstruct~\citep{yue2023mammoth} (a composite of $\sim$100k mathematical problems drawn from sources including GSM8K~\citep{cobbe2021training} and MATH~\citep{hendrycksmath2021}); we do not mix the two datasets in a single run. The SlimOrca-trained models are used for the capacity-scaling and training-dynamics analyses of Sections~\ref{subsec:scaling}--\ref{subsec:training_dynamics}, while the MathInstruct-trained models are used for the downstream accuracy evaluation of Section~\ref{subsec:downstream_accuracy}; the spectral analysis in Section~\ref{sec:mechanism} reports both. For evaluation, we use 500 competition-level MATH problems\footnote{\label{fn:math500-algebra}We take the first 500 problems of the \texttt{MATH-lighteval} test split in dataset order. Because that split is grouped by subject, these 500 problems all fall under the \emph{Algebra} category, spanning difficulty Levels 1--5. Our MATH pass@1, therefore, measures competition-level \emph{algebraic} reasoning specifically and should not be confused with the subject-stratified MATH-500 subset of \citet{lightman2024let}. We use the same fixed 500-problem slice for every method, so all reported comparisons remain matched.} and the full GSM8K test set (1,319 problems). MATH and GSM8K pass@1 are computed with greedy decoding, while MATH pass@10 uses nucleus sampling ($T = 0.8$, top-$p = 0.95$) with $10$ samples per problem and is reported with the unbiased pass@$k$ estimator of~\citet{chen2021evaluating}. The exact match is computed by extracting the last \verb|\boxed{}| expression from the generation (falling back to regex patterns when absent) and comparing it to the gold answer after LaTeX normalization---stripping \verb|\text|/\verb|\textbf|-style wrappers, canonicalizing \verb|\frac|, and removing whitespace. All methods are evaluated under identical settings.

We compare CeRA with LoRA and the SOTA linear variant DoRA~\citep{liu2024dora}. Evaluation metrics include Perplexity (PPL) for predictive performance, exact match (pass@k) for downstream problem-solving accuracy, and Effective Rank (ER) for spectral utilization.

\subsection{The Capacity Scaling Behavior}
\label{subsec:scaling}

We first evaluate whether non-linearity mitigates the rank saturation observed with linear adapters on the SlimOrca dataset with respect to predictive performance.

\begin{figure}[tb]
    \centering
    \includegraphics[width=0.48\columnwidth]{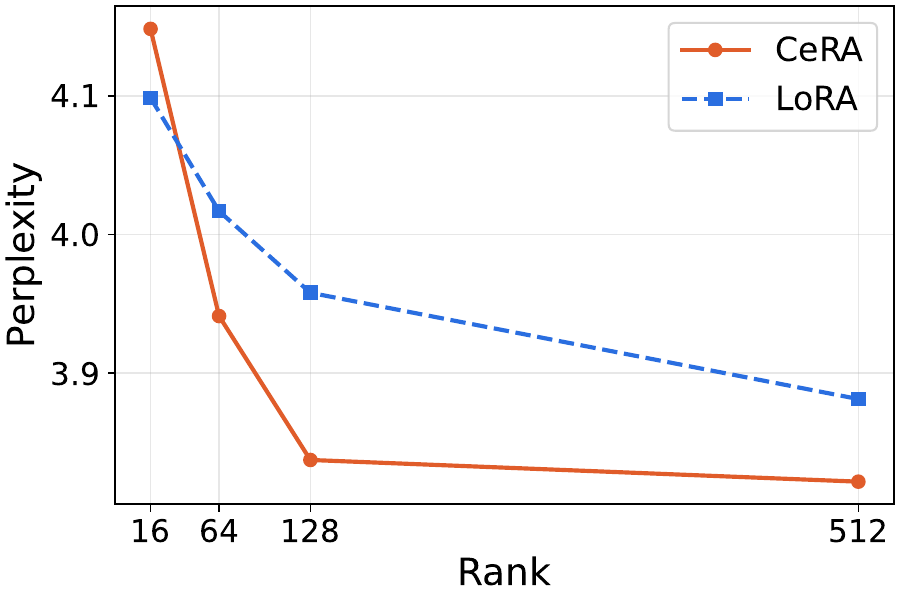}
    \caption{The capacity scaling behavior on SlimOrca. LoRA's performance plateaus rapidly, while CeRA ($\text{dropout}=0.1$) continues to improve (lower PPL is better).}
    \label{fig:orca_scaling}
\end{figure}

\paragraph{Perplexity Saturation.} 

Figure~\ref{fig:orca_scaling} illustrates the rapid diminishing returns for LoRA. Increasing the parameter budget by $32\times$ (from $r=16$ to $r=512$) yields a performance plateau around a perplexity of 3.90. This is consistent with the linear ceiling: additional ranks yield minimal gains in expressivity. In contrast, CeRA reaches a perplexity of 3.84 at rank 128, outperforming LoRA at rank 512 ($\approx 3.88$) with $4\times$ fewer parameters; even at rank 64 (3.94), CeRA already matches LoRA at rank 128. This indicates that the plateau is a structural bottleneck rather than a parameter-capacity issue.

\subsection{Optimization Robustness to Suboptimal Hyperparameters}
\label{subsec:training_dynamics}

A critical yet often overlooked bottleneck in deploying PEFT is optimization fragility. Linear adapters like LoRA require exhaustive hyperparameter searches because their performance is sensitive to the learning rate. To evaluate whether CeRA's capacity expansion mitigates this fragility, we analyze training dynamics under different learning rate regimes.

\begin{figure}[tb]
    \centering
    \includegraphics[width=\columnwidth]{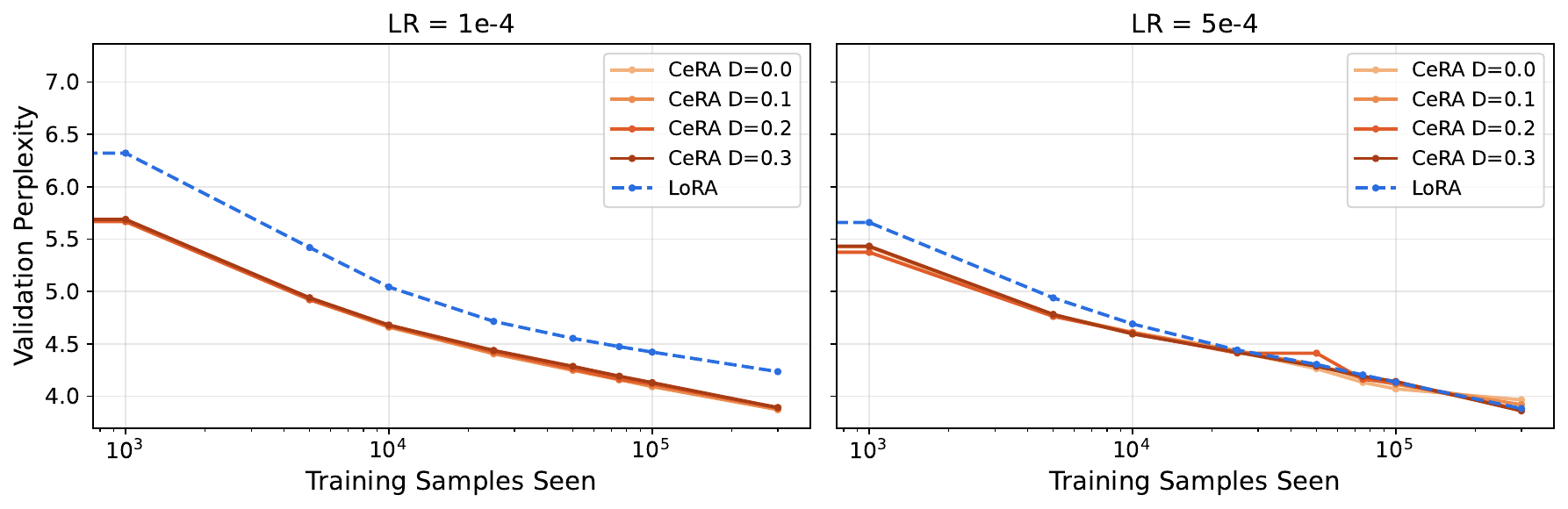}
    \caption{Validation perplexity during training on SlimOrca at a low (LR=1e-4) and a higher (LR=5e-4) learning rate, for CeRA at dropout $D\in\{0.0,0.1,0.2,0.3\}$ and LoRA without dropout. At the low learning rate, LoRA's descent slows and a perplexity gap persists, whereas CeRA descends smoothly to a competitive final perplexity; at the higher learning rate the methods converge comparably.}
    \label{fig:convergence_dynamics}
\end{figure}

Figure~\ref{fig:convergence_dynamics} shows the validation perplexity trajectories on SlimOrca for CeRA at dropout $D\in\{0.0,0.1,0.2,0.3\}$ and for LoRA (which uses no dropout), at a lower ($1\times10^{-4}$) and a higher ($5\times10^{-4}$) learning rate.

\paragraph{Comparable convergence at the higher learning rate.}
In the right panel ($5\times10^{-4}$), a learning rate that suits the linear baseline, both LoRA and CeRA variants converge aggressively and reach comparable early-stage plateaus. Adding nonlinear structural expansion therefore does not slow convergence or harm stability in this regime.

\paragraph{Robustness to a low learning rate.}
In the left panel ($1\times10^{-4}$), LoRA's descent slows and a perplexity gap persists throughout training, whereas every CeRA configuration continues to descend smoothly to a lower final perplexity. CeRA thus tolerates an under-tuned (low) learning rate better than LoRA on this objective. We do not claim uniform learning rate robustness: at the opposite extreme---the largest rank under the highest learning rate---CeRA itself degrades, as the complete learning rate sweep in Appendix~\ref{app:hyperparameters} shows.

\subsection{Downstream Task Accuracy and the Role of Task Complexity}
\label{subsec:downstream_accuracy}

Although PPL measures the predictive distribution, it is crucial to verify whether CeRA yields correct final answers. We evaluate exact-match (pass@k) accuracy on two datasets of contrasting difficulty: GSM8K (fundamental arithmetic) and MATH (complex mathematics).

\begin{table}[tb]
    \centering
    \caption{Downstream task accuracy (exact match). CeRA ($r=64$) attains the highest MATH pass@1, MATH pass@10, and GSM8K pass@1 of all configurations, exceeding or matching DoRA ($r=64$, $r=128$) and high-rank LoRA ($r=512$) at one-eighth of its parameter budget.}
    \label{tab:math_downstream}
    \begin{tabular}{lccccc}
      \toprule
      Method & Rank & Params & \multicolumn{2}{c}{MATH (Complex)} & GSM8K (Basic) \\
      & & & pass@1 & pass@10 & pass@1 \\
      \midrule
      LoRA
      & 64  & 27.3M  & 21.6\%          & 54.8\%          & 54.44\% \\
      & 128 & 54.5M  & 20.6\%          & 51.2\%          & 55.57\% \\
      & 512 & 218.1M & \underline{22.4\%} & \underline{55.4\%} & \underline{57.16\%} \\
      \midrule
      DoRA
      & 64  & 27.3M  & 19.8\%          & 54.6\%          & 55.12\% \\
      & 128 & 54.5M  & 21.2\%          & 51.4\%          & 51.63\% 
      \\
      \midrule
      CeRA (Ours)
      & 64  & 27.3M  & \textbf{23.6\%} & \textbf{56.6\%} & \textbf{58.76\%}
      \\
      \bottomrule
    \end{tabular}
\end{table}

\paragraph{High Parameter Efficiency.}

Table~\ref{tab:math_downstream} shows that CeRA ($r=64$) reaches a pass@1 of 23.6\% on MATH, the highest of all configurations, exceeding LoRA of the same rank 64 (21.6\%) and matching the accuracy of an $8\times$ larger LoRA at $r=512$ (22.4\%). By introducing non-linearity, CeRA overcomes the linear ceiling without inflating the parameter budget.

On the high-complexity MATH dataset, DoRA at rank 128 (21.2\%) outperforms LoRA at the same rank (20.6\%) but remains below CeRA at rank 64 (23.6\%). This indicates that altering the adapter's functional class via inference-time non-linearity is, on MATH in our setup, a more effective strategy than optimizing the gradient dynamics within a linear subspace.

For each method and rank, we select the learning rate on MATH pass@1 and report the corresponding test accuracy; the full sweep is shown in Table~\ref{tab:lr_sweep} in Appendix~\ref{app:hyperparameters}.

\paragraph{Capacity Expansion vs. Linear Re-parameterization.}
We compare CeRA against DoRA, a SOTA method that decomposes weight updates into magnitude and direction. On the high-complexity MATH dataset, DoRA improves over LoRA at the matched rank 128 but remains bound by its linear formulation. CeRA at rank 64 surpasses DoRA at the same rank, and, crucially, CeRA achieves this with half of DoRA's rank-128 budget, so the gain reflects the expanded functional class rather than added parameters. The case study below traces how the gap between linear and nonlinear adapters surfaces on a single reasoning problem, using LoRA as the linear representative.

\paragraph{Task Complexity Comparison.}
Comparing results across datasets reveals a relationship between adapter expressivity and task complexity. On the basic GSM8K dataset, the intrinsic dimensionality is low; LoRA's linear subspace suffices. DoRA's lower GSM8K accuracy at rank 128 in our runs is consistent with the observation that LoRA variants favor different learning rate ranges, so a learning rate selected for one objective need not be optimal for another~\citep{lee2026learning}; we report each method at its MATH-pass@1-selected learning rate (Appendix~\ref{app:hyperparameters}).

\subsection{Qualitative Analysis: Escaping the Linear Trap}
\label{subsec:qualitative_analysis}

Although quantitative metrics provide solid evidence of CeRA's superiority, it is crucial to examine how capacity expansion translates into actual reasoning capabilities. We conduct qualitative case studies. This section shows an iterative reasoning task. Iterative problems are challenging because they require the model to maintain and update a dynamic hidden state across multiple time steps without incurring error accumulation or state collapse.

\begin{table}[tb]
    \centering
    \caption{A case study on iterative reasoning. LoRA with a larger rank of 512 suffers from state collapse. In contrast, CeRA, despite its smaller rank, successfully models nonlinear recursive updates, generating a dynamic trajectory with far fewer parameters.}
    \label{tab:case_study}
    \small
    \begin{tabular}{@{}p{\linewidth}@{}}
        \toprule
        Q: Consider the logistic map $x_{n+1} = 3.5 x_n (1 - x_n)$ with $x_0 = 0.4$. Calculate the first few iterations. \\
        \midrule
        LoRA (Rank 512) -- State Collapse \\
        Step 1: $x_1 = 3.5(0.4)(0.6) = 0.84$. \\
        Step 2: $x_2 = 3.5(0.84)(0.16) = 0.4704$. \\
        Step 3: $x_3 = 3.5(0.4704)(0.5296) \approx 0.8719$. \\
        Step 4: $x_4 = 0.8719$. \\
        Step 5: $x_5 = 0.8719$. \\
        ... (Model repeats 0.8719 indefinitely) \\
        \midrule
        CeRA (Rank 128) -- Dynamic Tracking \\
        Step 1: $x_1 = 3.5(0.4)(0.6) = 0.84$. \\
        Step 2: $x_2 = 3.5(0.84)(0.16) = 0.4704$. \\
        Step 3: $x_3 = 3.5(0.4704)(0.5296) \approx 0.8719$. \\
        Step 4: $x_4 = 3.5(0.8719)(0.1281) \approx 0.3909$. \\
        Step 5: $x_5 = 3.5(0.3909)(0.6091) \approx 0.8333$. \\
        (Model continues to update values dynamically) \\
        \bottomrule
    \end{tabular}
\end{table}

Table~\ref{tab:case_study} presents a representative example. LoRA ($r=512$) exhibits state collapse. After correctly calculating the first three steps, the model fails to update its internal representation for the next iteration. It degenerates into a loop and outputs the same value at every following step, suggesting that LoRA's linear space is perhaps too rigid to capture the hidden dynamics of the recursive function. In contrast, CeRA, even with a smaller rank, recognizes that the value must change at each step. Although CeRA uses a smaller rank budget, its nonlinear components enable the model to project hidden states into a broader representation space, thereby escaping the linear trap.

Additional cases in both directions---including problems where LoRA's extra capacity does help, such as complex-number arithmetic---are collected in Appendix~\ref{app:cases}.

\section{Mechanism \& Design Analysis}
\label{sec:mechanism}

\subsection{Spectral Signature, Effective Rank, \& Manifold Dimensionality}
\label{subsec:spectral}

We analyze the singular-value spectrum of the learned adapters and quantify the structural expansion using the effective rank (ER) metric~\citep{roy2007effective}. The same signature appears in both SlimOrca and MathInstruct, the latter providing a mechanistic correlate of CeRA's MATH accuracy (Section~\ref{subsec:downstream_accuracy}).

\paragraph{Effective Rank.}
The effective rank measures the actual dimensionality of the information encoded in the activation space. For a matrix of adapter activations $H$ with normalized singular values $p_i$, the effective rank is the exponential of the Shannon entropy. We mean-center the activation matrix $H$ (column-wise) before computing its singular values.
\begin{equation}
    \text{ER}(H) = \exp \left( - \sum_{i=1}^k p_i \ln p_i \right)
\end{equation}
A higher ER implies a more uniform distribution of energy across dimensions, indicating a broader representation space.

\paragraph{Manifold dimensionality.} We also report the manifold dimensionality $d_{90}$, the smallest number of leading singular directions whose cumulative variance reaches 90\% of the total:
\begin{equation}
    d_{90}(H) = \min\Big\{ k : \frac{\sum_{i=1}^{k}\sigma_i^2}{\sum_{j}\sigma_j^2} \ge 0.9 \Big\},
\end{equation}
where $\sigma_1 \ge \sigma_2 \ge \cdots$ are the singular values of the (mean-centered) activation matrix $H$. ER weights every direction by its entropy; $d_{90}$ is a hard threshold for explained variance; the two are complementary measures of how broadly variance is spread across the spectrum.

\begin{figure}[tb]
    \centering
    \includegraphics[width=0.32\columnwidth]{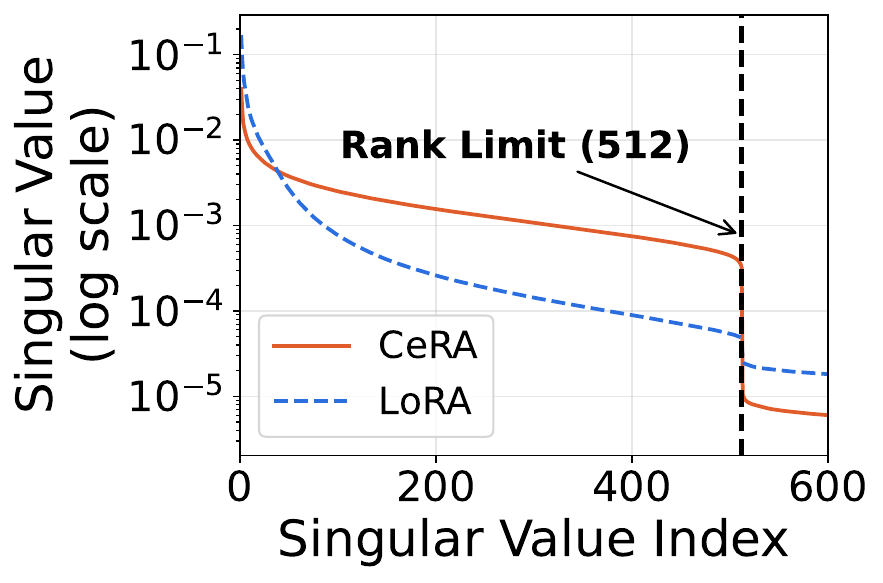}
    \hfill
    \includegraphics[width=0.32\columnwidth]{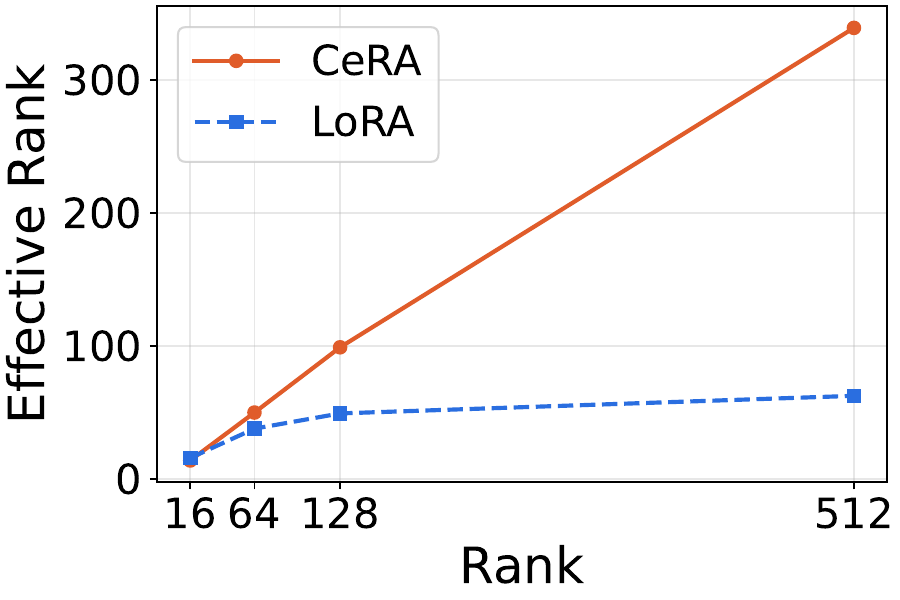}
    \hfill
    \includegraphics[width=0.32\columnwidth]{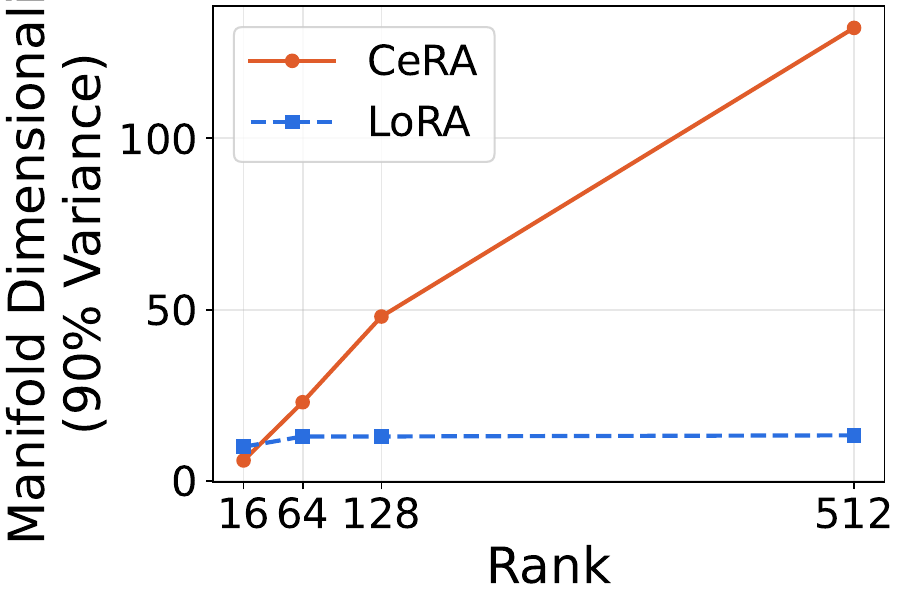}
    \caption{Spectral Analysis on SlimOrca. Left: Spectral Signature. LoRA exhibits rank collapse, whereas CeRA maintains a heavy tail. Middle: Effective Rank scales efficiently for CeRA but saturates for LoRA. Right: manifold dimensionality (90\% variance). LoRA plateaus early, whereas CeRA's spectral dimensionality keeps growing with rank.}
    \label{fig:orca_svd_spectrum}
\end{figure}

\begin{figure}[tb]
    \centering
    \includegraphics[width=0.32\columnwidth]{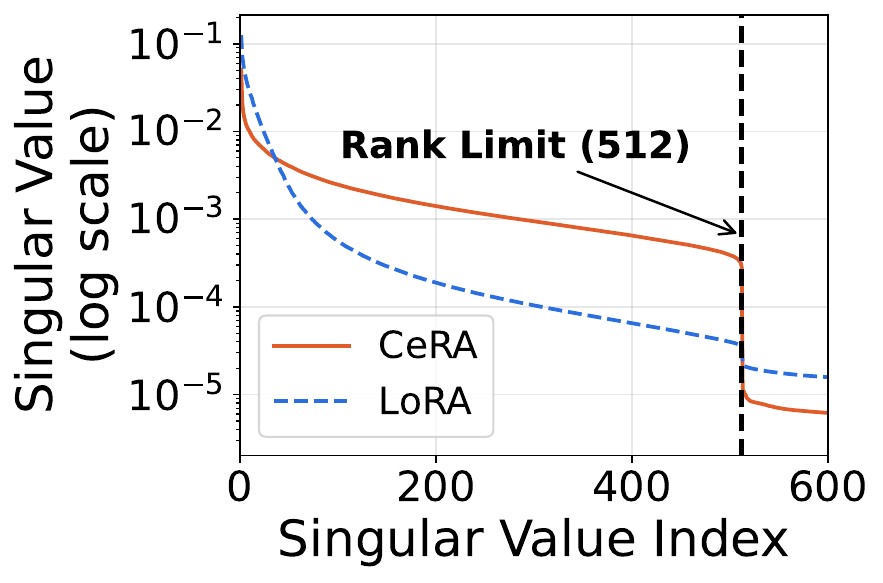}
    \hfill
    \includegraphics[width=0.32\columnwidth]{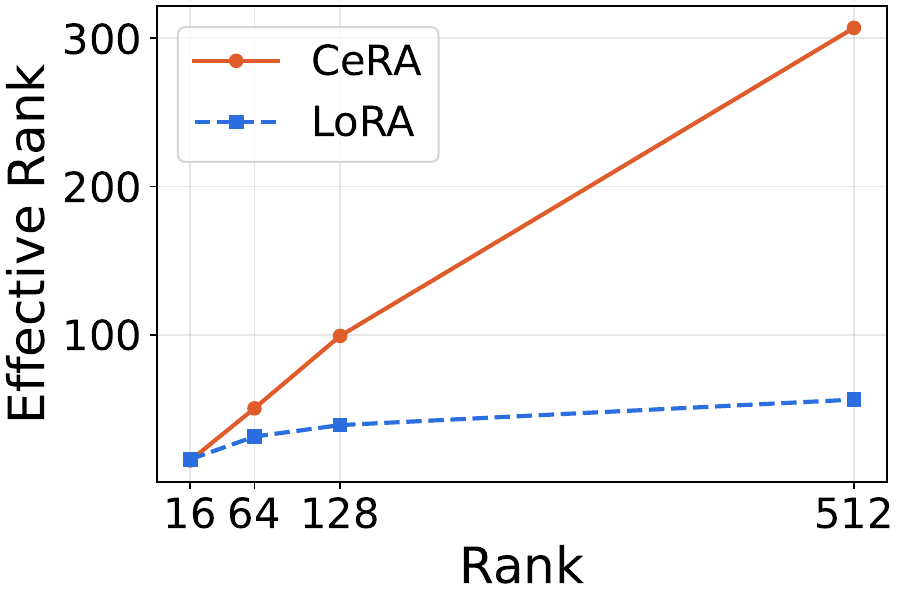}
    \hfill
    \includegraphics[width=0.32\columnwidth]{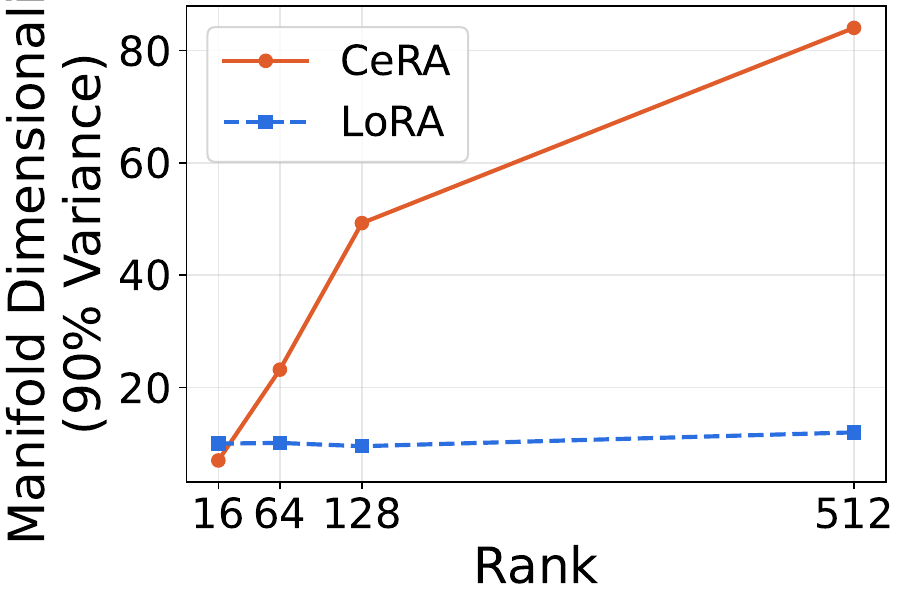}
    \caption{Spectral Analysis on MathInstruct. Left: Spectral Signature. LoRA exhibits a sharp drop. Middle: Effective Rank. LoRA's capacity saturates early. Right: Manifold Dimensionality (90\% variance). CeRA requires more singular components to reach 90\%.}
    \label{fig:math_spectral_analysis}
\end{figure}

\paragraph{Spectral signature and effective rank across datasets.}
On both datasets, LoRA exhibits rank collapse: its singular values decay sharply after the leading directions (Figures~\ref{fig:orca_svd_spectrum} and~\ref{fig:math_spectral_analysis}, left), so increasing the rank budget contributes little to the expressivity. CeRA instead retains energy in the tail of the spectrum and expands its effective rank with the budget (Figure~\ref{fig:orca_svd_spectrum}, middle). The same contrast holds on MathInstruct (Figure~\ref{fig:math_spectral_analysis}, middle): LoRA's effective rank saturates early while CeRA's keeps growing; CeRA's learned update does not collapse into a narrow low-rank subspace.

\paragraph{Manifold dimensionality and its link to reasoning.}
The manifold dimensionality---the number of singular directions needed to capture 90\% of the variance---tells the same story (Figures~\ref{fig:orca_svd_spectrum} and~\ref{fig:math_spectral_analysis}, right): LoRA plateaus early, whereas CeRA's dimensionality grows with rank: the optimizer recruits the tail dimensions rather than concentrating variance in a few dominant directions. On MathInstruct, this broader representation coincides with CeRA's higher MATH pass@1 accuracy (Section~\ref{subsec:downstream_accuracy}).

\paragraph{Interpretation of the spectral results.}
These spectral contrasts are descriptive, not causal. The linear baselines and CeRA train under different effective output scales ($\alpha/r$ vs.\ $s=1$), and a control experiment that matches the scale (Appendix~\ref{app:er_control}) shows that output scale, not non-linearity, accounts for most of the ER difference between LoRA and CeRA; moreover, ER does not track task performance (Section~\ref{subsec:ablation}). We therefore use ER and $d_{90}$ as summaries of the learned update's spectrum, and ground CeRA's justification in downstream accuracy and perplexity.

\subsection{Ablation Study} \label{subsec:ablation}

\begin{table}[tb]
    \centering
    \caption{Deconstructing CeRA (rank 512). We report ER and test PPL on SlimOrca. ER here is computed per adapted module and then averaged; its absolute values are therefore not directly comparable to the pooled-spectrum ER of Table~\ref{tab:er_control} in Appendix~\ref{app:er_control}, and comparisons are meaningful only within this table.}
    \label{tab:ablation}
    \begin{tabular}{llcc}
        \toprule
        Model Variant & Component Change & ER $\uparrow$ & Test PPL $\downarrow$ \\ \midrule
        CeRA & Full Architecture (SiLU, Dropout 0.3) & \textbf{376.90} & 3.91 \\
        \midrule
        (a) Granularity & Module-level (Parallel) & 354.52 & 3.94 \\
        \midrule
        (b) Activation & Identity (Linear) & 354.11 & 4.24 \\
         & ReLU & 339.70 & \textbf{3.69} \\
        \midrule
        (c) Dropout & No Dropout & 375.23 & 3.99 \\
        \bottomrule
    \end{tabular}
\end{table}

We isolate CeRA's core components on SlimOrca. We run the ablation at rank 512 rather than a smaller rank because, at low rank, the ER of all variants saturates quickly and their differences compress, whereas at rank 512, the spectrum has room to separate them (Table~\ref{tab:ablation}).

\paragraph{Granularity.} Replacing the weight-level with a module-level parallel adapter degrades PPL and lowers ER, confirming that intervening inside the $W_q$/$W_v$ projections matters.

\paragraph{Activation.} Here, PPL and ER diverge. Identity, which removes the activation entirely, is the most damaging change to predictive performance (PPL $4.24$). Both non-linearities we test lower PPL relative to this linear variant, but they behave oppositely on the spectrum: ReLU attains the lowest PPL ($3.69$) yet collapses ER to $339.70$---below even the linear baseline ($354.11$)---whereas SiLU expands it (to $376.90$). The spectral expansion is thus specific to SiLU, not a generic consequence of adding non-linearity. Because our downstream MATH gains are not predicted by SlimOrca PPL alone, we justify the SiLU activation on downstream accuracy rather than validation perplexity.

\paragraph{Dropout.} Removing structural dropout raises PPL but leaves ER essentially unchanged (375.23 vs.~376.90). Dropout therefore contributes mainly as a regularizer of the bottleneck, not as a driver of spectral expansion; the spectral expansion is attributable to SiLU.

\subsection{Hyperparameter Robustness and Optimization Stability}
\label{subsec:robustness}

A key factor for the adoption of a PEFT method is its robustness to hyperparameter choices. Recent variants, while effective in certain regimes, often introduce fragility in optimization. We evaluate the learning rate sensitivity of CeRA relative to the weight-decomposed DoRA.

\paragraph{Optimization Fragility in Linear Re-parameterization.}
DoRA decouples magnitude and direction to enhance flexibility. More generally, \citet{lee2026learning} show that a LoRA variant's forward design and update rule can reshape the loss Hessian over training, and that the usable learning rate range scales inversely with sharpness (the largest Hessian eigenvalue). We read DoRA's rank-dependent learning rate window in this light, consistent with the oscillations we observe in its high learning rate region.

\paragraph{Plug-and-Play Stability of CeRA.}
Because CeRA expands capacity through non-linearity rather than re-parameterizing the base-weight gradient pathways, its optimization landscape is generally stable across the learning-rate range we test. In our experiments, CeRA reaches competitive or better convergence under the same hyperparameter suite as LoRA at small and medium ranks; the one exception is the largest rank under aggressive learning rates ($r = 512$; see Section~\ref{subsec:training_dynamics} and Appendix~\ref{app:hyperparameters}), where CeRA degrades.

\subsection{Computational and Memory Complexity}
\label{subsec:complexity}

Since CeRA introduces non-linearity, the adapter weights cannot be merged into the base weights ($W_0 + \Delta W$) for zero-latency inference. We provide an analysis of the computational and memory overhead incurred by this unmerged paradigm to demonstrate its viability.

\paragraph{FLOPs Overhead Analysis.}
Let a linear projection layer with input dimension $d$, output dimension $k$, and adapter rank $r$. An unmerged linear adapter requires an additional $O(d \times r + r \times k)$ FLOPs. For CeRA, the forward pass requires the same matrix multiplications as the unmerged linear adapter, plus the element-wise SiLU activation, which takes $O(r)$ operations. Since $r \ll \min(d, k)$, the term $O(r)$ is negligible. Thus, the theoretical FLOPs overhead of CeRA is nearly identical to that of an unmerged LoRA.

\paragraph{Empirical Throughput in Multi-Tenant Serving.} 
In multi-tenant serving infrastructures (e.g., S-LoRA~\citep{sheng2023slora}, Punica~\citep{chen2024punica}), dynamic adapter switching is required to serve diverse user requests. In these systems, all adapters must be evaluated in an unmerged state. Under this paradigm, our benchmarks with Llama-3.1-8B indicate that CeRA incurs only a 6\% latency overhead compared to the unmerged linear baseline. The throughput remains consistently stable across varying ranks ($\approx 51$ tokens/second for both $r=64$ and $r=128$). The marginal latency gap is dominated primarily by memory-bound kernel-launching costs for the SiLU operation, rather than by arithmetic bottlenecks. 

\paragraph{Memory Utilization.}
During training, CeRA's memory footprint is highly efficient. Because a low-rank CeRA ($r=64$) can achieve the performance of a high-rank LoRA ($r=512$), the optimizer states and gradient buffers from the trainable parameters are reduced by a factor of 8. This reduction in VRAM outweighs the extra activation memory to store the intermediate SiLU states. Since these nonlinear activations occur only within the heavily compressed $r$-dimensional latent space ($r \ll d$), their spatial overhead is marginal.

\section{Related Work}
\label{sec:related}

LoRA~\citep{hu2022lora} established the standard for PEFT, assuming that weight updates reside in a low-rank linear subspace. Subsequent methods optimize this paradigm through dynamic rank allocation (AdaLoRA~\citep{zhang2023adalora}), weight decomposition (DoRA~\citep{liu2024dora}), or weight quantization (QLoRA~\citep{dettmers2023qlora}). Although these methods improve learning dynamics, they remain bound by the linear formulation. CeRA is orthogonal to them; it alters the functional form of the update, replacing the linear subspace with a nonlinear capacity expansion. We further distinguish methods by \emph{when} their non-linearity acts: Type-1 methods are nonlinear only during training and reduce to a mergeable affine map at inference, whereas CeRA (Type-2) retains its non-linearity at inference and stays non-mergeable. Among nonlinear low-rank methods, the closest are LoRAN~\citep{li2024loran}, which is also Type-2, and AuroRA~\citep{dong2026aurora}, a Type-1 method whose non-linearity collapses to an affine map at inference. Concurrently, \citet{lee2026learning} report that mergeable linear LoRA variants rarely beat vanilla LoRA once learning rates are tuned; CeRA differs in kind by retaining a non-mergeable inference-time non-linearity, and our experiments accordingly compare all methods within matched (rank, learning-rate) cells (Appendices~\ref{app:hyperparameters} and~\ref{app:model-scale}). A detailed methodological comparison with nonlinear low-rank adapters is given in Appendix~\ref{app:nonlinear_peft}.

\begin{table}[tb]
    \centering
    \caption{Feature Comparison. Unlike LoRA and its variants, CeRA introduces nonlinearity. Unlike traditional adapters---both sequential (Houlsby) and parallel---which operate at the module level, CeRA operates at the fine-grained weight level, enabling precise capacity expansion within attention projections.}
    \label{tab:related_comparison}
    \begin{tabularx}{\textwidth}{@{}l|XXXX@{}}
    \toprule
        Feature & LoRA & Houlsby Adapter & Parallel Adapter & CeRA (Ours) \\ \midrule
        Insertion Point & Weight-level & Module-level & Module-level & Weight-level \\
        (Granularity) & ($W_q, W_v$ internal) & (After Attn/FFN) & (Parallel to Attn/FFN) & ($W_q, W_v$ internal) \\ \midrule
        Structure & Linear Update & Nonlinear MLP & Nonlinear MLP & Nonlinear bottleneck MLP (SiLU)  \\ \midrule
        Non-Linearity & No & Yes & Yes & Yes \\ \midrule
        Mergeable? & Yes & No & No & No \\ \midrule
        Expressivity & Low & High & High & High \\
        \bottomrule
    \end{tabularx}
\end{table}

Furthermore, unlike early sequential adapters~\citep{houlsby2019parameter} or traditional parallel adapters~\citep{he2021towards, zhu2021counter} that operate coarsely at the module level (processing the aggregate output of an attention block), CeRA operates at the fine-grained weight level. As summarized in Table~\ref{tab:related_comparison}, CeRA modifies the attention mechanism's internal feature dynamics by injecting non-linearity into the internal query and value projections.

\section{Conclusion and Limitations} \label{sec:conc_and_limitations}

We revisit the linear assumption in Parameter-Efficient Fine-Tuning. For reasoning-intensive tasks, we observe that increasing LoRA's rank yields diminishing returns, raising the question of whether linear adapters face a capacity ceiling. We introduce CeRA, a weight-level architecture that uses SiLU gating and structural dropout to induce nonlinear capacity expansion. Under a fixed parameter budget, CeRA is competitive with much larger linear adapters: at rank 64 it matches or exceeds LoRA and DoRA of the same rank and a rank-512 LoRA in exact-match accuracy on MATH, using one-eighth of the trainable parameters. On MATH in our setting, altering the adapter's structural capacity is thus a more effective use of budget than optimizing gradient dynamics within a linear subspace.

Although CeRA offers clear advantages, it has limitations. First, because CeRA retains its non-linearity at inference, its update cannot be merged into the base weights ($W_0 + \Delta W$) for zero-latency inference. In multi-tenant serving, this is not a drawback: many adapters share a single frozen base and are kept unmerged in any case---in the cloud (S-LoRA~\citep{sheng2023slora}, Punica~\citep{chen2024punica}) and on edge devices (EdgeLoRA~\citep{shen2025edgelora})---so the additional cost is modest. However, for single-adapter deployment under strict latency or memory-bandwidth budgets, CeRA's non-mergeability is a genuine cost, and an affine-at-inference, mergeable method (LoRA, or AuroRA's static form~\citep{dong2026aurora}) is the appropriate choice. Second, although we evaluate across model scales within the Llama family (1B, 3B, 8B; Appendix~\ref{app:model-scale}), our analysis is confined to the Llama-3 family. Validating the capacity-expansion hypothesis on other architectures (e.g., Mixture-of-Experts) and exploring hybrid designs that pair DoRA's stable optimization with CeRA's expressivity remain future work. Third, our linear and nonlinear adapters do not share a matched effective gain: the linear baselines use a fixed $\alpha=32$, so their update scale $\alpha/r$ \emph{decreases} with rank, whereas CeRA uses unit scaling. This regime may suppress high-rank adapters~\citep{kalajdzievski2023rank}, so the rank saturation we interpret as a linear ceiling may be partly due to scaling.

\section*{Acknowledgments and GenAI Usage Disclosure}
We acknowledge support from the National Science and Technology Council of Taiwan under grant number 113-2221-E-008-100-MY3. The authors used Gemini and Claude to improve language and readability. The authors used Claude Code to help with the coding and experimentation. The authors reviewed and edited the content and code as needed.

\bibliographystyle{plainnat}  
\bibliography{ref}  

\appendix
\section{Hyperparameter Choice and Complete Learning Rate Sweep} \label{app:hyperparameters}

To ensure a fair comparison, we conducted an extensive learning rate sweep for all evaluated methods (LoRA, DoRA, and CeRA) under identical training budgets. Rather than reporting a single tuned configuration per method, we present the full sweep so that comparisons can be made within identical (rank, learning rate) settings. The dropout rate was set to $p=0.1$ for all CeRA main results, while the LoRA and DoRA baselines use no dropout. The higher $p=0.3$ appears only in the SlimOrca ablation (Table~\ref{tab:ablation}).

\paragraph{Optimization Setup.}
All models were fine-tuned with the AdamW optimizer, using bfloat16 precision and gradient clipping to a max norm of 1.0. Batch size (4), gradient accumulation steps (16, i.e., an effective batch size of 64), and sequence length (512) were kept constant across all methods to control for batch-level optimization effects.

\paragraph{Sweep Protocol.}
For each method, we sweep the learning rate over $\{1\times10^{-4}, 3\times10^{-4}, 5\times10^{-4}, 1\times10^{-3}\}$ at ranks $\{64, 128, 512\}$, with $\alpha = 32$. The full grid is reported in Table~\ref{tab:lr_sweep}. The best learning rate per (method, rank) is selected on MATH pass@1; this selection is applied identically to all methods and yields the values in Table~\ref{tab:math_downstream}. We use this complete grid, rather than only the per-method best configurations, for the matched-cell analysis below.

\begin{table}[ht]
\centering
\caption{Complete learning rate sweep on MATH pass@1 (\%) for Llama-3.1-8B. Each row is a matched (rank, learning rate) setting; \textbf{bold} marks the best learning rate per (method, rank), i.e.\ the values reported in Table~\ref{tab:math_downstream}. \checkmark\ marks cells where CeRA matches or exceeds the baseline at that identical setting: CeRA $\geq$ LoRA in 10 of 12 cells and CeRA $\geq$ DoRA in 10 of 12. The $r=512$/$5\text{e-}4$ cell ties with DoRA; all CeRA shortfalls relative to LoRA occur at $r=512$ under the highest learning rates.}
\label{tab:lr_sweep}
\begin{tabular}{cc rrr cc}
\toprule
Rank & LR & LoRA & DoRA & CeRA & CeRA $\geq$ LoRA & CeRA $\geq$ DoRA \\
\midrule
\multirow{4}{*}{64}
 & $1\times10^{-4}$ & 12.6 & 9.8  & 14.6 & \checkmark & \checkmark \\
 & $3\times10^{-4}$ & \textbf{21.6} & \textbf{19.8} & \textbf{23.6} & \checkmark & \checkmark \\
 & $5\times10^{-4}$ & 18.2 & 18.6 & 22.0 & \checkmark & \checkmark \\
 & $1\times10^{-3}$ & 15.0 & 7.2  & 21.8 & \checkmark & \checkmark \\
\midrule
\multirow{4}{*}{128}
 & $1\times10^{-4}$ & 13.4 & 11.2 & 15.6 & \checkmark & \checkmark \\
 & $3\times10^{-4}$ & \textbf{20.6} & 15.8 & 21.6 & \checkmark & \checkmark \\
 & $5\times10^{-4}$ & 17.0 & \textbf{21.2} & 18.0 & \checkmark & \\
 & $1\times10^{-3}$ & 13.4 & 18.2 & \textbf{22.0} & \checkmark & \checkmark \\
\midrule
\multirow{4}{*}{512}
 & $1\times10^{-4}$ & 14.4 & 12.6 & 19.6 & \checkmark & \checkmark \\
 & $3\times10^{-4}$ & \textbf{22.4} & 16.6 & \textbf{23.0} & \checkmark & \checkmark \\
 & $5\times10^{-4}$ & 18.4 & 17.0 & 17.0 &  & \checkmark \\
 & $1\times10^{-3}$ & 18.0 & \textbf{17.6} & 10.6 &  &  \\
\bottomrule
\end{tabular}
\end{table}

\paragraph{Matched-cell Comparison (CeRA vs.\ LoRA and CeRA vs.\ DoRA).}
Because cross-run comparisons at mismatched learning rates can be misleading, we compare CeRA vs.~LoRA (and CeRA vs.~DoRA) only within identical (rank, learning rate) settings. As shown in Table~\ref{tab:lr_sweep}, CeRA matches or exceeds LoRA in 10 of the 12 matched cells; under a one-sided sign test, this consistency is significant at $p\approx0.019$. The two exceptions both occur at the largest rank ($r=512$) under the two highest learning rates, where CeRA's accuracy drops. This matched-cell consistency, rather than any single headline gap, is the basis for our claim that inference-time non-linearity---not parameter count---is what lifts the linear ceiling; the same holds against DoRA (CeRA matches or wins in 10 of 12). We emphasize this because \citet{lee2026learning} show that mergeable, linear LoRA variants (e.g., PiSSA, MiLoRA, DoRA) do not consistently beat vanilla LoRA once their learning rates are tuned, with peak accuracies typically within $1$--$2\%$. CeRA lies outside the class they study: it retains an input-dependent non-linearity at inference (Type-2, non-mergeable). Our evidence is correspondingly not a single tuned gap but (i) the matched-cell consistency above (Table~\ref{tab:lr_sweep}) and (ii) the rank-scaling perplexity margin across 1B/3B/8B (Appendix~\ref{app:model-scale}), both controlled exactly for learning rate.

\paragraph{LoRA.}
Vanilla LoRA attains its best MATH pass@1 at $3\times10^{-4}$ across all ranks, which we therefore use as its reference configuration. Its accuracy degrades steadily as the learning rate exceeds this threshold, reflecting the learning rate sensitivity commonly reported for linear adapters.

\paragraph{DoRA.}
DoRA is the most sensitive to the learning rate of the three methods. It underfits at the lowest learning rate ($1\times10^{-4}$: $9.8$/$11.2$/$12.6$ across ranks) and becomes unstable at the highest ($1\times10^{-3}$: $7.2$ at $r=64$). Its best learning rate is not uniform but varies with rank ($3\times10^{-4}$ at $r=64$, $5\times10^{-4}$ at $r=128$). This rank-dependent usable window is consistent with the general sharpness--learning-rate relationship of~\citet{lee2026learning}. We also empirically observe DoRA's relative weakness on GSM8K (Table~\ref{tab:math_downstream}).

\paragraph{CeRA.}
CeRA reaches its best or near-best accuracy at $3\times10^{-4}$---the same learning rate that is optimal for LoRA---so no separate tuning was required to switch from LoRA to CeRA. CeRA is notably robust on the high learning rate side at small and medium rank: at $r=64$ it stays within $\sim 2$ points across $\{3\text{e-}4, 5\text{e-}4, 1\text{e-}3\}$ ($23.6$/$22.0$/$21.8$), and at $r=128$ it does not collapse at $1\times10^{-3}$ ($22.0$). This robustness has two boundaries we state explicitly: like all methods, CeRA underfits at the lowest learning rate ($1\times10^{-4}$), and at the largest rank ($r=512$), its accuracy drops under aggressive learning rates ($10.6$ at $1\times10^{-3}$). 

\section{Model Scaling Test} \label{app:model-scale}

\begin{table}[t]
\centering
\caption{Llama-3.2-1B, MathInstruct held-out perplexity (lower is better), from the same runs as Table~\ref{tab:scale_1b}. \textbf{Bold} denotes the best learning rate for each (method, rank). \checkmark\ marks cells where CeRA $\leq$ the baseline; CeRA is lower than both LoRA and DoRA in all 12 matched cells, and the margin widens with rank.}
\label{tab:scale_1b_ppl}
\begin{tabular}{cc rrr cc}
\toprule
Rank & LR & LoRA & DoRA & CeRA & CeRA $\leq$ LoRA & CeRA $\leq$ DoRA \\
\midrule
\multirow{4}{*}{64}
 & $1\times10^{-4}$ & 2.62 & 2.62 & 2.60 & \checkmark & \checkmark \\
 & $3\times10^{-4}$ & 2.48 & 2.48 & 2.46 & \checkmark & \checkmark \\
 & $5\times10^{-4}$ & 2.45 & \textbf{2.43} & \textbf{2.41} & \checkmark & \checkmark \\
 & $1\times10^{-3}$ & \textbf{2.44} & \textbf{2.43} & \textbf{2.41} & \checkmark & \checkmark \\
\midrule
\multirow{4}{*}{128}
 & $1\times10^{-4}$ & 2.62 & 2.62 & 2.54 & \checkmark & \checkmark \\
 & $3\times10^{-4}$ & 2.47 & 2.46 & 2.38 & \checkmark & \checkmark \\
 & $5\times10^{-4}$ & 2.40 & \textbf{2.40} & \textbf{2.34} & \checkmark & \checkmark \\
 & $1\times10^{-3}$ & \textbf{2.39} & \textbf{2.40} & 2.36 & \checkmark & \checkmark \\
\midrule
\multirow{4}{*}{512}
 & $1\times10^{-4}$ & 2.62 & 2.62 & 2.39 & \checkmark & \checkmark \\
 & $3\times10^{-4}$ & 2.46 & 2.46 & \textbf{2.24} & \checkmark & \checkmark \\
 & $5\times10^{-4}$ & 2.40 & 2.39 & \textbf{2.24} & \checkmark & \checkmark \\
 & $1\times10^{-3}$ & \textbf{2.36} & \textbf{2.35} & 2.29 & \checkmark & \checkmark \\
\bottomrule
\end{tabular}
\end{table}

\begin{table}[t]
\centering
\caption{Llama-3.2-3B, MathInstruct held-out perplexity, from the same runs as Table~\ref{tab:scale_3b}. Conventions as in Table~\ref{tab:scale_1b_ppl}; CeRA is lower than both baselines in all 12 matched cells.}
\label{tab:scale_3b_ppl}
\begin{tabular}{cc rrr cc}
\toprule
Rank & LR & LoRA & DoRA & CeRA & CeRA $\leq$ LoRA & CeRA $\leq$ DoRA \\
\midrule
\multirow{4}{*}{64}
 & $1\times10^{-4}$ & 2.34 & 2.34 & 2.32 & \checkmark & \checkmark \\
 & $3\times10^{-4}$ & 2.20 & 2.20 & 2.17 & \checkmark & \checkmark \\
 & $5\times10^{-4}$ & \textbf{2.14} & 2.15 & \textbf{2.11} & \checkmark & \checkmark \\
 & $1\times10^{-3}$ & \textbf{2.14} & \textbf{2.11} & \textbf{2.11} & \checkmark & \checkmark \\
\midrule
\multirow{4}{*}{128}
 & $1\times10^{-4}$ & 2.33 & 2.34 & 2.26 & \checkmark & \checkmark \\
 & $3\times10^{-4}$ & 2.19 & 2.18 & 2.08 & \checkmark & \checkmark \\
 & $5\times10^{-4}$ & 2.12 & 2.11 & \textbf{2.02} & \checkmark & \checkmark \\
 & $1\times10^{-3}$ & \textbf{2.09} & \textbf{2.09} & 2.06 & \checkmark & \checkmark \\
\midrule
\multirow{4}{*}{512}
 & $1\times10^{-4}$ & 2.34 & 2.34 & 2.09 & \checkmark & \checkmark \\
 & $3\times10^{-4}$ & 2.18 & 2.18 & \textbf{1.93} & \checkmark & \checkmark \\
 & $5\times10^{-4}$ & 2.09 & 2.10 & 1.94 & \checkmark & \checkmark \\
 & $1\times10^{-3}$ & \textbf{2.03} & \textbf{2.06} & 2.00 & \checkmark & \checkmark \\
\bottomrule
\end{tabular}
\end{table}

\begin{table}[t]
\centering
\caption{Llama-3.1-8B, MathInstruct held-out perplexity, from the same runs as Table~\ref{tab:lr_sweep}. Conventions as in Table~\ref{tab:scale_1b_ppl}. CeRA is lower than both LoRA and DoRA in 11 of 12 cells; the sole exception is the ``high rank/high learning rate'' collapse at $r=512$/$1\times10^{-3}$ (the same cell where MATH pass@1 collapses).}
\label{tab:scale_8b_ppl}
\begin{tabular}{cc rrr cc}
\toprule
Rank & LR & LoRA & DoRA & CeRA & CeRA $\leq$ LoRA & CeRA $\leq$ DoRA \\
\midrule
\multirow{4}{*}{64}
 & $1\times10^{-4}$ & 2.014 & 2.014 & 2.003 & \checkmark & \checkmark \\
 & $3\times10^{-4}$ & 1.940 & 1.939 & 1.906 & \checkmark & \checkmark \\
 & $5\times10^{-4}$ & \textbf{1.919} & \textbf{1.916} & \textbf{1.885} & \checkmark & \checkmark \\
 & $1\times10^{-3}$ & 1.956 & 1.952 & 1.899 & \checkmark & \checkmark \\
\midrule
\multirow{4}{*}{128}
 & $1\times10^{-4}$ & 2.013 & 2.016 & 1.963 & \checkmark & \checkmark \\
 & $3\times10^{-4}$ & 1.929 & 1.927 & 1.862 & \checkmark & \checkmark \\
 & $5\times10^{-4}$ & \textbf{1.916} & \textbf{1.904} & \textbf{1.828} & \checkmark & \checkmark \\
 & $1\times10^{-3}$ & 1.936 & 1.921 & 1.881 & \checkmark & \checkmark \\
\midrule
\multirow{4}{*}{512}
 & $1\times10^{-4}$ & 2.013 & 2.014 & 1.852 & \checkmark & \checkmark \\
 & $3\times10^{-4}$ & 1.910 & 1.915 & \textbf{1.790} & \checkmark & \checkmark \\
 & $5\times10^{-4}$ & 1.910 & \textbf{1.893} & 1.837 & \checkmark & \checkmark \\
 & $1\times10^{-3}$ & \textbf{1.897} & \textbf{1.893} & 2.417 &  &  \\
\bottomrule
\end{tabular}
\end{table}

\begin{table}[t]
\centering
\caption{Llama-3.2-1B, complete learning rate sweep (MATH pass@1, \%). Each row is a matched (rank, learning rate) setting; \textbf{bold} denotes the best learning rate for each (method, rank). \checkmark\ marks cells where CeRA $\geq$ the baseline at that identical setting: CeRA $\geq$ LoRA in all 12 cells and $\geq$ DoRA in 9 of 12. Absolute MATH accuracy is near the noise floor at this scale ($\leq 5\%$), so these consistencies are directional only.}
\label{tab:scale_1b}
\begin{tabular}{cc rrr cc}
\toprule
Rank & LR & LoRA & DoRA & CeRA & CeRA $\geq$ LoRA & CeRA $\geq$ DoRA \\
\midrule
\multirow{4}{*}{64}
 & $1\times10^{-4}$ & 0.8 & 1.6 & 1.6 & \checkmark & \checkmark \\
 & $3\times10^{-4}$ & \textbf{1.8} & 1.6 & \textbf{3.2} & \checkmark & \checkmark \\
 & $5\times10^{-4}$ & 1.6 & 2.4 & 1.6 & \checkmark & \\
 & $1\times10^{-3}$ & 1.2 & \textbf{3.2} & 2.2 & \checkmark & \\
\midrule
\multirow{4}{*}{128}
 & $1\times10^{-4}$ & 0.4 & 1.6 & 1.0 & \checkmark & \\
 & $3\times10^{-4}$ & 2.2 & 1.2 & \textbf{5.0} & \checkmark & \checkmark \\
 & $5\times10^{-4}$ & \textbf{3.0} & 2.0 & 3.2 & \checkmark & \checkmark \\
 & $1\times10^{-3}$ & 2.2 & \textbf{2.6} & 2.8 & \checkmark & \checkmark \\
\midrule
\multirow{4}{*}{512}
 & $1\times10^{-4}$ & 1.8 & 1.4 & 3.2 & \checkmark & \checkmark \\
 & $3\times10^{-4}$ & 2.6 & 1.8 & \textbf{4.6} & \checkmark & \checkmark \\
 & $5\times10^{-4}$ & \textbf{3.0} & \textbf{2.2} & 3.4 & \checkmark & \checkmark \\
 & $1\times10^{-3}$ & \textbf{3.0} & 2.0 & 4.0 & \checkmark & \checkmark \\
\bottomrule
\end{tabular}
\end{table}

\begin{table}[t]
\centering
\caption{Llama-3.2-3B, complete learning rate sweep (MATH pass@1, \%). Conventions as in Table~\ref{tab:scale_1b}. CeRA $\geq$ LoRA in 6 of 12 matched cells and $\geq$ DoRA in 8 of 12; on MATH, the three methods are statistically comparable at this scale.}
\label{tab:scale_3b}
\begin{tabular}{cc rrr cc}
\toprule
Rank & LR & LoRA & DoRA & CeRA & CeRA $\geq$ LoRA & CeRA $\geq$ DoRA \\
\midrule
\multirow{4}{*}{64}
 & $1\times10^{-4}$ & 7.4 & 5.4 & 7.0 &  & \checkmark \\
 & $3\times10^{-4}$ & \textbf{11.0} & 7.8 & \textbf{10.8} &  & \checkmark \\
 & $5\times10^{-4}$ & 8.0 & 8.2 & 10.4 & \checkmark & \checkmark \\
 & $1\times10^{-3}$ & 9.6 & \textbf{8.6} & 9.8 & \checkmark & \checkmark \\
\midrule
\multirow{4}{*}{128}
 & $1\times10^{-4}$ & 4.8 & 5.6 & 7.8 & \checkmark & \checkmark \\
 & $3\times10^{-4}$ & 8.4 & \textbf{11.0} & 7.4 &  & \\
 & $5\times10^{-4}$ & \textbf{11.2} & \textbf{11.0} & \textbf{8.8} &  & \\
 & $1\times10^{-3}$ & 9.0 & 9.8 & 8.4 &  & \\
\midrule
\multirow{4}{*}{512}
 & $1\times10^{-4}$ & 6.0 & 5.4 & 10.2 & \checkmark & \checkmark \\
 & $3\times10^{-4}$ & 8.2 & 8.8 & 7.8 &  & \\
 & $5\times10^{-4}$ & \textbf{10.0} & \textbf{11.2} & \textbf{11.2} & \checkmark & \checkmark \\
 & $1\times10^{-3}$ & 6.6 & 8.0 & 10.4 & \checkmark & \checkmark \\
\bottomrule
\end{tabular}
\end{table}

To test whether CeRA's behavior is specific to the 8B backbone, we repeat the rank and learning rate sweep on two smaller models in the same family, Llama-3.2-1B and Llama-3.2-3B, fine-tuned on MathInstruct with AdamW at a fixed learning rate. For each (rank, learning rate) cell, we report greedy pass@1 on our 500-problem algebra slice and MathInstruct held-out perplexity, computed from the same run. The 8B accuracy sweep appears in Appendix~\ref{app:hyperparameters} (Table~\ref{tab:lr_sweep}); its perplexity is in Table~\ref{tab:scale_8b_ppl}.

\paragraph{The fitting advantage is consistent and scales with rank.}
CeRA attains a lower MathInstruct held-out perplexity than \emph{both} LoRA and DoRA in 35 of the 36 matched (rank, learning rate) cells across 1B, 3B, and 8B (Tables~\ref{tab:scale_1b_ppl},~\ref{tab:scale_3b_ppl},~\ref{tab:scale_8b_ppl}). The sole exception is the ``high rank/high learning rate'' cell ($r{=}512$, $1\times10^{-3}$) at 8B, where CeRA's training destabilizes---the same cell in which its MATH pass@1 collapses (Table~\ref{tab:lr_sweep}), so the instability appears consistently across both metrics. Elsewhere, the margin widens with rank on every scale: the best learning rate gap over LoRA grows from $\approx$0.03 at $r=64$ to $\approx$0.10 -- 0.12 at $r=512$, mirroring the capacity-scaling trend observed in SlimOrca in Figure~\ref{fig:orca_scaling}. CeRA continues to use an additional rank where the linear baselines plateau.

\paragraph{Downstream accuracy is gated by backbone capability.}
The exact-match picture is scale-dependent (Tables~\ref{tab:scale_1b},~\ref{tab:scale_3b}). Despite CeRA's near-uniform perplexity advantage, the gain appears only as exact-match accuracy, where MATH is actually learnable. At 8B, MATH is the discriminating task and CeRA $\geq$ LoRA in 10 of 12 matched cells (Table~\ref{tab:lr_sweep}). At 1B, MATH sits near the noise floor for all methods ($\leq 5\%$ pass@1) and is uninformative as an accuracy benchmark. However, CeRA's direction is consistent (CeRA $\geq$ LoRA in all 12 cells, at very small absolute values). At 3B, MATH is only weakly learnable, and CeRA and LoRA are statistically comparable on it (CeRA $\geq$ LoRA in 6 of 12 cells), although CeRA's perplexity is lower (better) in all 12. Better distribution fitting, therefore, does not automatically convert into higher exact-match accuracy when the task sits at the backbone's capability frontier.

\paragraph{Interpretation.} 
Read together, these results indicate that CeRA's nonlinear capacity yields a consistent, rank-scaling improvement in the adapter's fit to the target distribution across all scales. Meanwhile, its \emph{downstream} advantage emerges only when the backbone is sufficiently capable to represent the harder task. This refines, rather than contradicts, the task-complexity relationship of Section~\ref{subsec:downstream_accuracy}: ``complexity'' is relative to backbone capability, not absolute.

\paragraph{Robustness.} 
CeRA's high learning rate stability persists at smaller scales: at $r\leq128$ it shows no collapse at $1\times10^{-3}$ on either backbone, whereas LoRA degrades at high learning rate (e.g., 3B MATH pass@1 $10.0\to6.6$ from $5\times10^{-4}$ to $1\times10^{-3}$). The high-rank collapse observed for CeRA appears only at 8B and $r=512$ under the highest learning rate.

\section{Extended Comparison with Nonlinear PEFT Methods} \label{app:nonlinear_peft}

\begin{table}[t]
\centering
\small
\caption{Nonlinear (NL) low-rank adaptation methods classified by whether the adapter's forward map is nonlinear \emph{in the input $x$} at training and at inference. Only methods that remain nonlinear in $x$ at inference (Type-2) are non-mergeable. CeRA is Type-2.}
\label{tab:nonlinear_peft}
\begin{tabular}{l l c c c l}
\toprule
Method & Mechanism & \makecell{NL in $x$\\(train)} & \makecell{NL in $x$\\(infer)} & Mergeable & Class \\
\midrule
LoRA         & Linear $BA$                         & No  & No  & Yes & Linear \\
DoRA        & Magnitude--direction reparam.       & No  & No  & Yes & Linear \\
PEANuT (NEAT) & Weight-space $f(W_0)$               & No  & No  & Yes & Linear \\
\midrule
AuroRA   & $\sigma(Ax)$ (train) $\to$ $\sigma(A)x$ (infer) & Yes & No  & Yes & Type-1 \\
\midrule
LoRAN  & $f(Ax)$, Sinter (sine)        & Yes & Yes & No  & Type-2 \\
CoLA$^{\dagger}$  & Autoencoder (pre-training)    & Yes & Yes & n/a & --- \\
\textbf{CeRA (Ours)}           & $\sigma(Ax)$, SiLU + dropout & Yes & Yes & No  & Type-2 \\
\bottomrule
\end{tabular}
\\[2pt]
\raggedright\footnotesize
$^{\dagger}$ CoLA pre-trains from scratch, not a PEFT adapter on a frozen base; mergeability axis does not apply.
\end{table}

We classify nonlinear low-rank adapters by whether the adapter's forward map is nonlinear in the input $x$ and, if so, whether that nonlinearity survives at inference (Table~\ref{tab:nonlinear_peft}). The first group is effectively \emph{linear}: LoRA, DoRA, and PEANuT (formerly NEAT)~\citep{zhong2026peanut} are linear in $x$ both in training and inference and remain mergeable. Although PEANuT applies a nonlinear network to the \emph{frozen} weights, $f(W_0)$, this yields a fixed update matrix, so the adapter stays linear in $x$. Its nonlinearity lies in the weight-space parameterization, not in the input--output map. \emph{Type-1} methods are nonlinear in $x$ during training but collapse to a fixed affine map at inference, so they remain mergeable: AuroRA~\citep{dong2026aurora} trains with an input-dependent update $B\sigma(Ax)$ but reverts at inference to a static update $\Delta W = B\sigma(A)$, which is affine in $x$. \emph{Type-2} methods retain their input-dependent nonlinearity at inference and are therefore not mergeable. CeRA is Type-2. Its closest neighbor is LoRAN~\citep{li2024loran}, which inserts a sinusoidal activation after the low-rank projection. CeRA differs in that it applies SiLU gating with structural dropout at the weight level (within $W_q$ and $W_v$). Finally, CoLA~\citep{liu2025cola} is a from-scratch pre-training method that replaces dense projections with low-rank autoencoders, targeting a different setting from PEFT on a frozen base, and therefore lies outside this taxonomy.

\section{Effective Rank vs.~Output Scale vs.~Non-linearity}
\label{app:er_control}

In the main paper, we report effective rank (ER) as a spectral diagnostic. We had initially intended ER to serve a stronger, mechanistic role---non-linearity expands the spectrum of the adapted representation---but a control experiment leads us to withdraw that causal reading. Holding the measurement pipeline, rank ($512$), learning rate ($5\times10^{-4}$), and dropout ($0$) fixed, we vary only the adapter's output scale and its activation (Table~\ref{tab:er_control}).

\begin{table}[tbh]
\centering
\caption{Effective rank under a fixed measurement pipeline ($r=512$, $\text{lr}=5\times10^{-4}$, dropout $=0$). ER is computed on the same activation matrix for all rows: a \emph{single} SVD over the pooled spectrum of all adapted modules. Table~\ref{tab:ablation} instead averages per-module ERs (and its identity variant uses dropout $0.3$), so absolute ER values are not comparable across the two tables; only within-table comparisons are meaningful. Raising the linear adapter's output scaling \emph{alone} accounts for most of the gap to CeRA; adding SiLU contributes only a small remainder.}
\label{tab:er_control}
\begin{tabular}{llc}
\toprule
Configuration & Scaling & ER \\
\midrule
LoRA ($\alpha=32$, default) & $0.0625$ & $100.11$ \\
LoRA ($\alpha=512$)         & $1.0$    & $299.60$ \\
CeRA, identity activation (linear) & $1.0$ & $348.62$ \\
CeRA, SiLU                  & $1.0$    & $387.56$ \\
\bottomrule
\end{tabular}
\end{table}

Two observations follow. First, output scale, not non-linearity, drives most of the ER difference. Raising LoRA's scaling from its default $\alpha/r = 0.0625$ to $1.0$ moves ER from $100.1$ to $299.6$ \emph{without any architectural change}; a linear CeRA variant (identity activation, unit scaling) already reaches $348.6$, and adding SiLU contributes only a further ${\sim}11\%$ ($387.6$). Of the total gap between default LoRA and CeRA-SiLU, roughly $70\%$ is attributable to scaling and $\sim\!14\%$ to the SiLU non-linearity. Because ER is invariant to a global rescaling of the activation matrix by construction (rescaling all singular values by a constant leaves the normalized spectrum, and hence the entropy, unchanged), this shift is not a measurement artifact; it reflects the fact that the training-time gain alters the \emph{learned} weights. The dependence on output scale also connects to a known property of the $\alpha/r$ convention, under which the effective update of a linear adapter is suppressed as rank grows~\citep{kalajdzievski2023rank}.

Second, ER does not track task performance. In our ablation (Table~\ref{tab:ablation}), the ReLU variant attains the \emph{lowest} ER yet the \emph{best} perplexity, while the linear variant has high ER but the \emph{worst} perplexity. ER is therefore neither a clean function of architecture nor a reliable predictor of downstream quality. We consequently use ER and $d_{90}$ as descriptive summaries of the adapted spectrum and ground CeRA's justification in downstream accuracy and perplexity rather than in spectral expansion. Isolating what genuinely distinguishes nonlinear from linear adapters once output scale is matched---for instance, via gain-matched spectral probes or conditional-linear-map analyses---remains open for future work.

\section{Case studies} \label{app:cases}

\begin{table*}[t]
\centering
\caption{Representative bidirectional cases, CeRA ($r=64$, 27.3M params) vs.\ LoRA ($r=512$, 218.1M, an $8\times$ larger budget) on Llama-3.1-8B, greedy pass@1. Top block: CeRA correct, LoRA wrong---here LoRA's failures are typically \emph{strategic or conceptual} (a wrong method or a misread structure). Bottom block: LoRA correct, CeRA wrong---here CeRA's failures are typically \emph{low-level execution slips} (a dropped sign, factor, or term). These cases are selected to illustrate that tendency, not absolute.}
\label{tab:appendix_cases}
\small
\renewcommand{\arraystretch}{1.2}
\begin{tabularx}{\textwidth}{@{}l X c c c X@{}}
\toprule
\textbf{ID} & \textbf{Problem (abbreviated)} & \textbf{Gold} &
\textbf{CeRA} & \textbf{LoRA} & \textbf{Decisive difference (loser's error)}\\
\midrule
\multicolumn{6}{@{}l}{\textit{CeRA $r=64$ correct \;/\; LoRA $r=512$ wrong --- LoRA errs strategically}}\\
\midrule
M14 & Kite area from vertex coordinates ($\tfrac12 d_1 d_2$) & $75$ & $75$ & $65$ & LoRA invents a false identity $(AB)(CD)=(AC)(BD)$ instead of using the diagonals.\\
M28 & Roots $a,b$ of $x^2-5x+9=0$; find $(a-1)(b-1)$ & $5$ & $5$ & --- & LoRA takes the quadratic formula into $\sqrt{-11}$ and never recovers; CeRA uses Vieta's formulas. \\
M87 & $-4$ is a root of $x^2+bx-36=0$; find $b$ & $-5$ & $-5$ & $8$ & LoRA misapplies Vieta, assuming \emph{both} roots equal $-4$.\\
M124 & $b{-}a$ for integer solutions of $x^2{-}15<2x$ & $6$ & $6$ & $-2$ & Both factor $(x{-}5)(x{+}3){<}0$; LoRA selects the \emph{complementary} region ($x{<}{-}3$ or $x{>}5$ instead of $-3{<}x{<}5$), so its endpoints are wrong.\\
G24 & \$19.50 after a 25\% discount; original price & $26$ & $26$ & $24.4$ & LoRA adds 25\% of the \emph{sale} price instead of dividing by $0.75$.\\
G30 & Ages $7:11$, total $162$; Allen's age in 10\,yr & $109$ & $109$ & $172$ & LoRA mis-applies the ratio, lands on Allen $=162$, and reports $162+10=172$; CeRA solves $18x=162$\\
G47 & $2\times$ red vs.\ blue ties, red 50\% pricier; total spend & $800$ & $800$ & $600$ & LoRA ignores the 50\% markup and mis-structures the counts.\\
\midrule
\multicolumn{6}{@{}l}{\textit{LoRA $r=512$ correct \;/\; CeRA $r=64$ wrong --- CeRA errs by a slip}}\\
\midrule
M47 & Simplify $(2-2i)(5+5i)$ & $20$ & $20i$ & $20$ & CeRA drops the sign on $-2i$ (expands as if $(2{+}2i)$): both cross terms become $+10i$ (so $+20i$, not $0$) and $+10i^2$ replaces $-10i^2$, collapsing the real part to $0$. Uses $i^2{=}{-}1$ correctly; result $20i$.\\
M59 & Simplify $(3-i)(6+2i)$ & $20$ & $16+6i$ & $20$ & CeRA mishandles the $i^2$ sign and keeps a stray $6i$.\\
M272 & Compute $(34{-}10){+}(20{-}9){+}(55{-}10)$ & $80$ & $100$ & $80$ & CeRA regroups correctly to $34{+}20{+}26$ but miscomputes the final sum as $100$ instead of $80$.\\
G18 & 3-egg omelet daily for 4 weeks; dozens of eggs & $7$ & $2$ & $7$ & CeRA drops the $\times3$ eggs/omelet factor (28 vs.\ 84 eggs).\\
G58 & \$40 bill, $+25\%$ fee, $+\$3$ delivery, $+\$4$ tip; total & $57$ & $47$ & $57$ & CeRA loses the \$10 ($25\%$) fee mid-calculation.\\
G60 & 25 oranges: 1 bad, 20\% unripe, 2 sour; good ones & $17$ & $18$ & $17$ & CeRA forgets to subtract the single bad orange.\\
G88 & Marilyn sold $10 \times$ Harald, 88,000 total; Harald's sales & $8000$ & $9777$ & $8000$ & CeRA writes $10x=88000-x$ then combines to $9x=88000$ (subtracts $x$ instead of adding), giving $88000/9 \approx 9777$.\\
\bottomrule
\end{tabularx}
\end{table*}

To see how CeRA's capacity expansion plays out problem by problem, we compare CeRA ($r=64$, 27.3M trainable parameters) with the strongest baseline from Table~\ref{tab:math_downstream} (LoRA, $r=512$, 218.1M, an $8\times$ larger budget). Both use greedy decoding, and both are evaluated on the same 500 MATH problems and the full GSM8K test set. We focus on \emph{disagreements}---problems that exactly one of the two adapters answers correctly---since these isolate behavioral differences that the aggregate accuracies blur.

\paragraph{Disagreement statistics.}
On MATH, the two models diverge on $112$ of $500$ problems: CeRA alone is correct on $59$ and LoRA on $53$ (a further $59$ are solved by both). On GSM8K they diverge on $353$ of $1,319$: CeRA alone is correct on $187$, LoRA alone on $166$. The near-symmetry of the raw counts is itself informative---the headline MATH gap ($23.6$ vs.\ $22.4$) understates how differently the two adapters reason and shows that CeRA's edge does not come from dominating a fixed subset of problems but from a different error profile achieved with one-eighth of the parameters. Table~\ref{tab:appendix_cases} collects representative cases from each direction.

\paragraph{An asymmetry in error type.}
Reading the disagreements, the two error profiles differ in kind, not just in count. When LoRA loses, it tends to fail at the level of \emph{strategy or structure}: it reaches for the quadratic formula and strands itself in complex roots where a Vieta's-formulas shortcut applies (M28), invents a false area identity rather than using the kite's diagonals (M14), selects the wrong region of a quadratic inequality (M124), or mishandles the structure of a percentage or ratio word problem (G24, G30, G47). When CeRA loses, by contrast, its high-level approach is usually sound, and the failure is a \emph{low-level execution slip}: a dropped sign in complex-number multiplication (M47, M59), a dropped multiplicative factor or term (G18, G58), an off-by-one (G60), or a plain arithmetic mistake (M272, G88). Consistent with the capacity argument, CeRA's wins also skew toward the hardest items: among the Level-5 algebra problems on which the two disagree, CeRA wins $11$ and LoRA $6$.

\paragraph{Scope and caveats.}
This asymmetry is a tendency, not a law, and we explicitly state its limits. It is the dominant pattern we observe, but counterexamples exist in both directions: LoRA also makes arithmetic slips, and---more importantly for our claim---CeRA also does fail \emph{conceptually}, for instance, by choosing a worse solution method (completing the square where factoring is clean) or setting up the wrong equation. The cases in Table~\ref{tab:appendix_cases} are selected to illustrate the prevailing trend, not to assert that CeRA never errs strategically. Therefore, we treat these traces as qualitative illustrations, with systematic evidence remaining from the matched-cell sweep (Appendix~\ref{app:hyperparameters}, Table~\ref{tab:lr_sweep}). A small number of raw disagreements are answer-extraction artifacts---a correct value rejected by the \verb|\boxed{}| parser because of trailing formatting---which we excluded when selecting examples.

\end{document}